\documentclass{article}
\usepackage{ijcai16}

\usepackage{times}

\usepackage[utf8]{inputenc}
\usepackage[american]{babel}
\usepackage{csquotes}
\usepackage[backend=bibtex,style=nature,doi=false,isbn=false,url=false,natbib=true]{biblatex}
\DeclareLanguageMapping{american}{american-apa} 

\usepackage{amsmath}
\usepackage{yfonts}
\usepackage{eufrak}
\usepackage{algpseudocode}
\usepackage{graphicx}
\usepackage{hyperref}
\usepackage{soul, color}
\usepackage{bbm}
\usepackage{subcaption}

\title{Toward a general, scaleable framework for Bayesian teaching with applications to topic models}  
\author{Baxter S. Eaves Jr \and Patrick Shafto \\ 
Rutgers Univerity - Newark  \\
\{baxter.eaves, patrick.shafto\}@rutgers.edu}

\begin{document}

\maketitle

\begin{abstract}
Machines, not humans, are the world's dominant knowledge accumulators but humans remain the dominant decision makers. Interpreting and disseminating the knowledge accumulated by machines requires expertise, time, and is prone to failure. The problem of how best to convey accumulated knowledge from computers to humans is a critical bottleneck in the broader application of machine learning. We propose an approach based on human teaching where the problem is formalized as selecting a small subset of the data that will, with high probability, lead the human user to the correct inference. This approach, though successful for modeling human learning in simple laboratory experiments, has failed to achieve broader relevance due to challenges in formulating general and scalable algorithms. We propose general-purpose teaching via pseudo-marginal sampling and demonstrate the algorithm by teaching topic models. Simulation results show our sampling-based approach: effectively approximates the probability where ground-truth is possible via enumeration, results in data that are markedly different from those expected by random sampling, and speeds learning especially for small amounts of data. Application to movie synopsis data illustrates differences between teaching and random sampling for teaching distributions and specific topics, and demonstrates gains in scalability and applicability to real-world problems.
\end{abstract}

\section{Introduction}

Machines are increasingly integral to society. Where expert human intuition once was, algorithms are increasingly present. 
Learning algorithms are used to learn complex hypotheses from complex data, which are used to augment human intuition. This paradigm creates a bottleneck where the knowledge accumulated by machines depends on highly trained human experts to interpret, and to conveying to humans. This remains a barrier to the broader usefulness of machine learning. While machines can communicate perfectly among themselves by exchanging bits, humans communicate with data---they teach. The purposeful selection of data plays a featured role in theories of cognition \citep{sperber1986relevance}, cognitive development \citep{Gergely2007}, and culture \citep{Tomasello2005}. In each of these cases, teaching is conceived of as purposeful, rather than random, selection of small set of examples, with the goal of facilitating accurate inferences about a body of knowledge. 

The  question of how to model teaching has also appeared across literatures: cognitive scientists have proposed and investigated probabilistic models of optimal example selection \citep{Shafto2008,Shafto2012,Shafto2014,Tenenbaum2001}; algorithmic teaching researchers have proposed deterministic methods to select examples that rule out confusable concepts \citep{balbach2008,zilleslhz2008}; and machine learning researchers have become interested in machine teaching \citep{Zhu2013b,Patil2014}. These paths of research vary in the details, but share the same goal: to facilitate learning by optimizing input to learners.

Previous approaches are either tailored to a specific, circumscribed domains or propose a general mathematical formalization that is not scalable.
\citet{Zhu2013b} offers a formalization for exponential family distributions \citep[see also][]{Tenenbaum2001}. \citet{Mei2015,Mei2015a} cast teaching as bilevel optimization, which is NP-hard, and offer a solution for the limited case when the data are differentiable and the learner's objective function is convex and regular. \citet{Shafto2008} and \citet{zilleslhz2008} offer general formalizations for probabilistic and deterministic inference, respectively, but suffer from computational complexity that renders them inapplicable for real-world application.

We propose a simple, general framework for selecting examples to teach probabilistic learners. The framework leverages advances in sampling-based approximations, specifically pseudo-marginal sampling coupled with importance sampling, to offer a general purpose approximation to the Bayesian normalizing constant that is required to teach probabilistic learners. We demonstrate the efficacy of the approach on probabilistic topic models. The results show that the distribution of teaching data differ significantly from the data likelihood, that the teaching data improve learning especially when learning from a small numbers of examples, and that teaching can be optimized for a variety of tasks, while scaling to problems markedly more complex than possible under previous, general-purpose, teaching algorithms. 

\section{Background}
\subsection{Bayesian teaching}
A teacher generates data, $x$, to lead a learner to a specific hypothesis, $\lambda$. A teacher must consider the learner's posterior inference, $\pi_L(\lambda | x)$, given every possible choice of data, thus learning is a sub-problem of teaching. The teacher simulates a learner who considers the probabilities of all hypotheses given that data,
\begin{equation}
    \label{eqn:teaching}
    p_{T}(x \mid \lambda) 
        = 
            \frac{
                \pi_{L}(\lambda \mid x)
            }{
                \int_x \pi_{L}(\lambda \mid x) dx
            }
        \propto
            \frac{
                \mathcal{\ell}_{L}(x \mid \lambda)
            }{
                m_{L}(x)
            }.
\end{equation}

\noindent
Where the subscripts $T$ and $L$ indicate probabilities attached to teacher and learner, respectively; $\mathcal{\ell}(x\mid\lambda)$ indicates the liklelihood of the data; and $m(x) = \int \mathcal{\ell}p(x,\lambda)d\lambda$ is the marginal likelihood.

\subsection{Topic models via Latent Dirichlet Allocation}

Latent Dirichlet Allocation \citep[LDA; ][]{Blei2003} is a popular formalization of topic models. Under LDA, documents are bags of words generated by mixtures of a fixed number topics. To generate a set of $D$ documents from $T$ topics with $W$ distinct words:
\begin{algorithmic}
    \ForAll {topics, $1, \dots, T$}
    \State $\phi_t \sim \text{Dirichlet}_W(\beta)$
    \EndFor

    \ForAll {documents, $1, \dots, D$}
    \State $\theta_d \sim \text{Dirichlet}_T(\alpha)$
        \For {$i \in \{1,\dots,W_d\}$}
            \State $z_i^{(d)} \sim \text{Categorical}( \theta_d )$
            \State $w_i^{(d)} \sim \text{Categorical}( \phi_{z_i}^{(d)} )$
        \EndFor
    \EndFor
\end{algorithmic}
\noindent
where $W_d$ is the number of words in the document $d$, $z_i^{(d)}$ is the topic from which the $i^{th}$ word in document $d$ was generated, and $w_i^{(d)}$ is the $i^{th}$ word in document $d$. The variables of interest are the topics, $\Phi = \{\phi_i,\dots,\phi_T\}$ and the topic mixture weights $\Theta = \{\theta_1,\dots,\theta_D\}$. Each $\phi$ is a vector with an entry for each of the $W$ words in the vocabulary; the $w^{th}$ entry is the probability of word $w$ occurring in the topic. Each document is associated with a $\theta$. The $t^{th}$ entry of $\theta$ is the probability that a word in document $d$ was generated from topic $t$.

The full joint distribution is,
\begin{equation*}
    p(\mathbf{z}, \mathbf{w}, \Phi, \Theta | \alpha, \beta)
    = 
    p(\mathbf{z} | \Theta)p(\mathbf{w} | \Phi)p(\Phi | \beta)p(\Theta | \alpha).
\end{equation*}
\noindent Exploiting conjugacy, we can integrate out $\phi$ and $\theta$ leaving,
\begin{equation*}
    p(\mathbf{z}, \mathbf{w} \mid T, \alpha, \beta)
    =
        \prod_{d=1}^{D} 
            \text{DirCat}(z^{(d)} | \alpha)
        \prod_{t=1}^{T} 
            \text{DirCat}(w^{(t)} | \beta),
\end{equation*}
where DirCat is the Dirichlet-Categorical distribution. Given a $k$-length vector of counts, $x$, 
\begin{equation*}
\text{DirCat}(x \mid \alpha) 
    =
        \frac{
            \Gamma \left( \sum_{i=1}^k \alpha_i \right)
        }{
            \Gamma \left( n + \sum_{i=1}^k \alpha_i \right)
        } 
        \prod\limits_{i=1}^k
            \frac{
                \Gamma \left( x_i + \alpha_i \right) 
            }{
                \Gamma \left( \alpha_i \right)
            }.
\end{equation*}
This allows one to define an efficient Gibbs sampler for $z$ that only requires maintaining a set of counts \citep{Griffiths2004}. The probability of word $i$ being assigned to a specific topic ($z_i=t$) given the words in the documents, $\mathbf{w}$, and the assignment of all other words, $\mathbf{z}^{(-i)}$, is,
\begin{equation}
    \label{eqn:lda-gibbs}
    p(z_i=t \mid \mathbf{z}^{(-i)}, \mathbf{w})
    \propto
    \left(
        n_{d,t}^{(-i)} + \alpha_d
    \right)
    \frac{
        n_{t,w}^{(-i)} + \beta_w
    }{
        \sum_{w'} n_{t,w}^{(-i)} + \beta_w
    },
\end{equation}
\noindent where $n_{d,t}^{(-i)}$ is the number of words---less word $i$---in document $d$ assigned to topic $t$, $n_{t,w}^{(-i)}$ is the number of instance of vocabulary word $w$---again, less word $i$---assigned to topic $t$.

\section{Teaching topics with documents}
Our goal is to produce or choose documents to teach a topic model, $\phi$, to an LDA learner. The learner is assumed to know the prior parameters, $\alpha$ and $\beta$, and the number of topics; but must marginalize over all possible assignments of $N$ words in $D$ documents to $T$ topics, $\mathbf{z} \in \mathfrak{Z}$, and all possible $\Theta$. The probability of documents under the teaching model is,
\begin{multline}
    \label{eqn:teaching-topics}
    p_T(\{d_1, \dots d_D\} | \Phi, \alpha, \beta)
    \propto 
        \prod_{\phi\in\Phi} f(\phi \mid \beta) \\ \times
        \sum_{\mathbf{z} \in \mathfrak{Z}}
            \left( 
                \left( \prod_{d=1}^{D}
                    \text{DirCat}(z_d \mid \alpha)
                \right)
                \prod_{i=1}^{n} 
                    f(w_i \mid \phi_{z_i})
            \right)
    \\ \div
    \sum_{\mathbf{z} \in \mathfrak{Z}}
        \left(
            \prod_{d=1}^{D} 
                \text{DirCat}(z_d \mid \alpha)
        \right)
        \left( 
            \prod_{t=1}^{T} 
                \text{DirCat}(w_t \mid \beta) 
        \right).
\end{multline}

\subsection{Approximating the LDA posterior}
\label{sec:approx}
Computing the above quantity requires $\mathcal{O}(T^N)$ computations; because in all but the simplest scenarios this is impossible, we must approximate it.

Importance sampling is a common approach to estimating intractable integrals (or sums) by re-framing them as expectations with respect to an easy-to-sample-from importance distribution, $q$. The integral of $p$ over $\lambda$ is estimated by simulating $\bar{\lambda}_1,\dots,\bar{\lambda}_M$ from $q(\lambda)$ and taking the arithmetic mean of the \emph{importance weights}, $p/q$. For example, to estimate the marginal likelihood,
\begin{equation*}
    m(x) = \int \mathcal{\ell}(x|\lambda)\pi(\lambda) d\lambda
         \approx \frac{1}{M}\sum_{i=1}^{M}\frac{\mathcal{\ell}(x|\bar{\lambda}_i)\pi(\bar{\lambda}_i)}{q(\bar{\lambda}_i)},\quad \bar{\lambda} \sim q.
\end{equation*}

When estimating the marginal likelihood, the importance distribution should be close to the posterior because areas with high posterior density contribute more. In the case of LDA, the naive approach of drawing $\mathbf{z}$ from a uniform categorical distribution is inefficient and does not afford scaling to real-world problems; many of the importance samples will come from low probability regions because the importance distribution will be far flatter than the target. A better approach is to use the collapsed Gibbs sampler in \autoref{eqn:lda-gibbs} to implement a \emph{sequential importance sampler} \citep[SIS; ][]{Maceachern1999} from which proposals are drawn by incrementally generating $\mathbf{z}$. Starting with a random value for $z_1$, draw $z_2 \mid z_1 , \{w_1, w_2\}, \alpha, \beta$, according to \autoref{eqn:lda-gibbs}, then draw $z_3 \mid \{z_1, z_2\} , \{w_1, w_2, w_3\}, \alpha, \beta$ and so on,
\begin{equation*}
    q(\mathbf{z}) = \prod_{i=2}^{n}p(z_i | \{z_1,\dots,z_{i-1}\}, \{w_1,\dots,w_{i}\}, \alpha, \beta).
\end{equation*}

\begin{figure}[t]
    \centering
    \includegraphics[trim={4.5mm 5mm 5mm 4.5mm},width=\columnwidth]{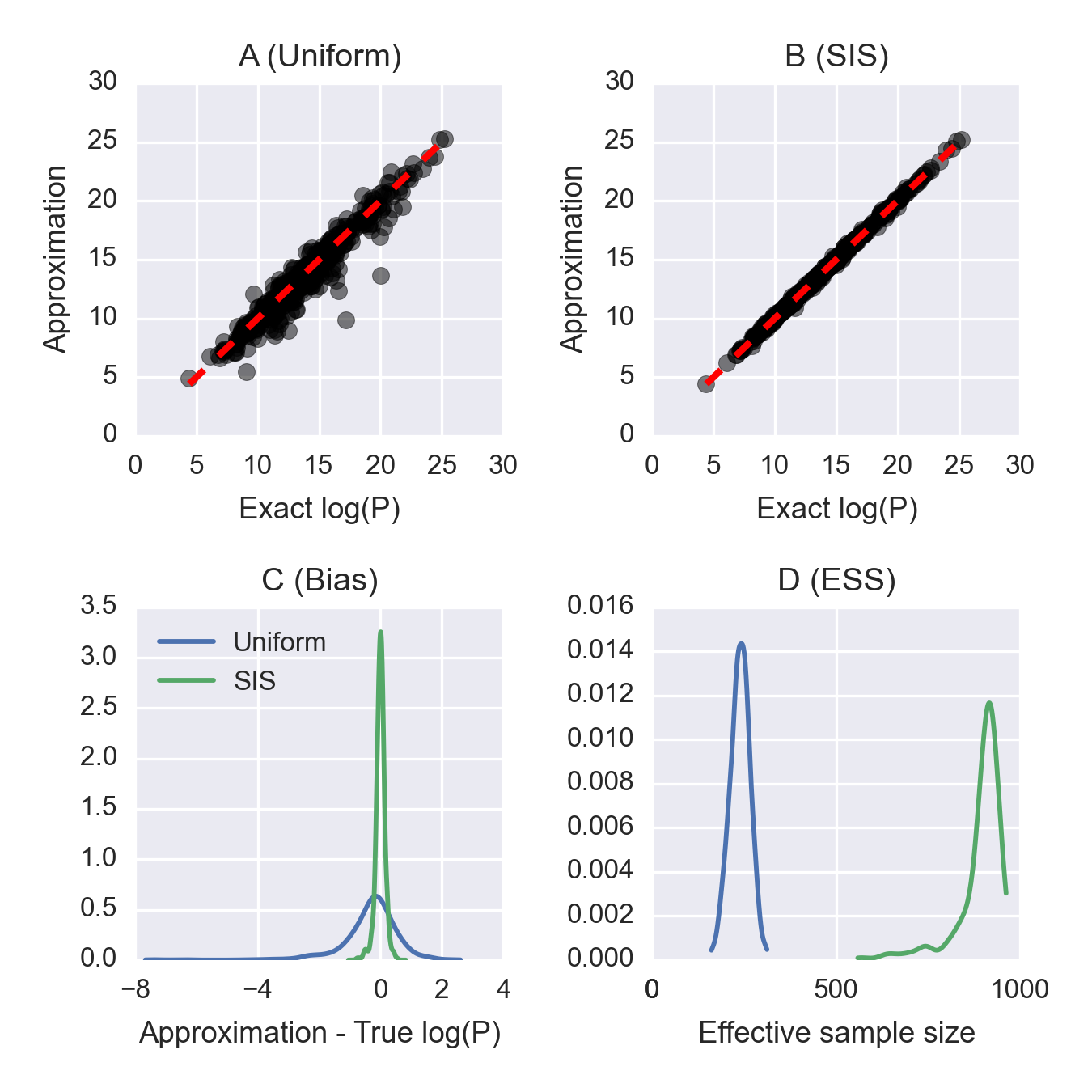}
    \caption{Comparison of teaching probability approximations. Panels A, B, and C show the true log teaching probabilities plotted against approximations from uniform importance sampling and sequential importance sampling (SIS). Each approximation was calculated using 1000 total samples. Each point on the plot represents the exact and approximate log probability for a pair of sampled documents with five words each generated from three topics with a five-word vocabulary ($\alpha = \beta = .5$). Panel C shows the bias of the log of the estimators. Panel D shows the effective sample sizes (ESS) of both of the estimators.}
    \label{fig:approx-compare}
\end{figure}

SIS performs significantly better than naive importance sampling, especially in sparser models (as $\alpha$ and $\beta$ approach zero). We evaluated the accuracy and efficiency of uniform and sequential importance sampling by comparing their estimates (after 1000 samples) against the true teaching probabilities. We generated 512 pairs of documents and three-topic models from LDA with $\alpha = \beta = .5$; each document contained five words from a five-word vocabulary. Given these parameters, the number of terms in the sum over $\mathfrak{Z}$ in \autoref{eqn:teaching-topics} is $3^{10} =  59049$. \autoref{fig:approx-compare}A, B, and C shows the accuracy of the samplers and indicates that SIS has much lower variability. \autoref{fig:approx-compare}D compares the \emph{effective sample size} \parentext{ESS, \parencite{Kong1992}} of the samplers. ESS represents the number of unweighted samples to which the $M$ weighted samples are equivalent---higher is better. ESS can be calculated as $M/(1 + \text{Var}_q(w))$, where $\text{Var}_q(w)$ is the sample variance of the importance weights. For these simulations, SIS had an average ESS of 890.71, while uniform importance sampling had an ESS of 238.50.

\subsection{Pseudo-marginal sampling}
To generate documents from the teaching distribution, we employ \emph{pseudo-marginal sampling} \citep{Andrieu2009, Andrieu2012}, which allows exact Metropolis-Hasting to be performed using approximated functions. The standard Metropolis-Hasting algorithm \citep{Metropolis1953,Hastings1970} generates samples, $y$, from a probability distribution, $p$, that is known to a constant, $p(y) = kf(y)$, by drawing new values, $y'$, from a proposal distribution, $q(y \rightarrow y')$, and accepting $y'$ with probability $\text{min}[1, A]$ where,
\begin{equation}
\label{eqn:mh}
A := \frac{f(y')q(y' \rightarrow y)}{f(y)q(y \rightarrow y')}.
\end{equation}

If $f$ is difficult to calculate, the pseudo-marginal approach allows one to replace $f$ in \autoref{eqn:mh} with an approximation $\hat{f}$. Pseudo-marginal sampling works provided $\hat{f}(y) = \gamma f(y)$, where $\gamma$ is the bias or weight of the approximation and the expected value of the weights is constant. The weights are implicitly treated as a random variable in a joint distribution and marginalized away, leaving the target. We will use pseudo-marginal sampling coupled with SIS to generate documents under the teaching model; when we do no need to generate documents, we use only approximation.

\section{Scalability}
\begin{figure}[htb!]
    \centering
    \includegraphics[trim={8mm 5mm 2mm 4mm}, clip, width=\columnwidth]{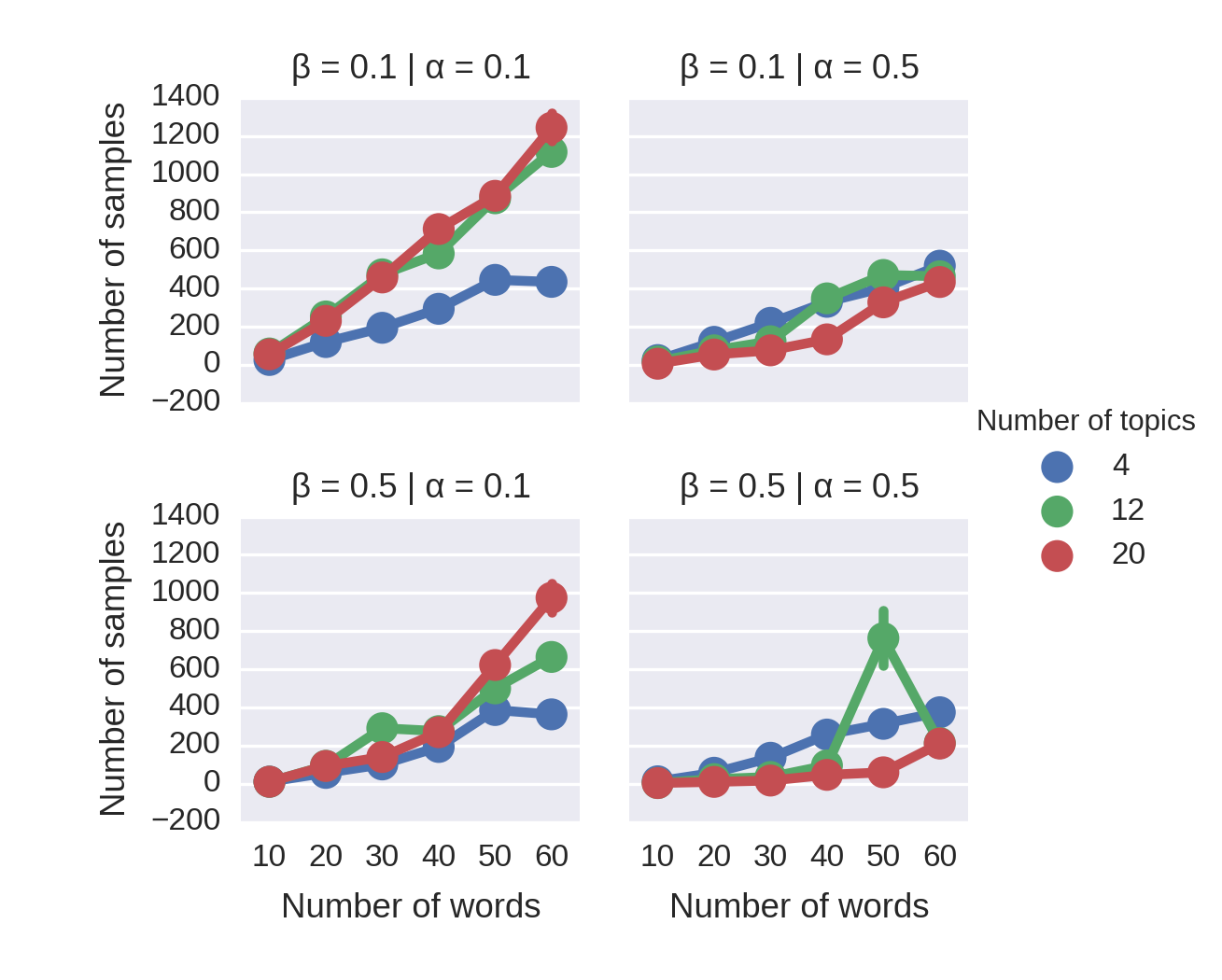}
    \caption{Number of samples to achieve 0.05 relative sample error when calculating the marginal likelihood for the teaching probability for single documents. Each point represents the average over 1024 runs. Error bars represent the 95\% confidence interval.}
    \label{fig:complexity}
\end{figure}

The posterior approximation is the main factor in scaling Bayesian teaching. For topic models, our chosen approximation method, SIS, requires only $\mathcal{O}(nT)$ arithmetic operations and $n$ multinomial random numbers \parentext{\autoref{eqn:lda-gibbs}}; and has been shown to yield high ESS. Because each sample is generated independently, importance sampling is embarrassingly parallel. However, as the number of words and topics increases, the number of samples needed to approximate the posterior within an acceptable error increases. 
We ran simulations to evaluate the number of SIS samples, $M$, required to achieve a relative sample error of 0.05 while calculating the marginal likelihood with different values of $n$, $T$, $\alpha$, and $\beta$, and with $W=100$. The relative error of the importance samples is,
\begin{equation}
    \frac{1}{\sqrt{M}}
    \sqrt{
        \frac{
            \frac{1}{M}\sum_{i=1}^{M} w_i^2
        }{
            \left(\frac{1}{M}\sum_{i=1}^{M} w_i\right)^2
        }-1
    },
\end{equation}
\noindent
where $w_i$ is an importance weight. \autoref{fig:complexity} displays the results, averaged over 1024 runs. The teaching probability of a single sixty-word document under a twenty-topic model can be calculated to within 0.05 relative sample error with about one thousand samples. The number of samples required increases linearly with the number of words. The values of $\alpha$ and $\beta$ effect scaling. As the model becomes sparser, a smaller proportion of assignments hold most of the probability mass, which makes high-mass areas more difficult for the sampler to find, thus more samples are required for sparser models.

\section{Examples}
\subsection{Characterizing the teaching distribution}
\begin{figure*}[ht!]
    \centering
    \includegraphics[width=\textwidth]{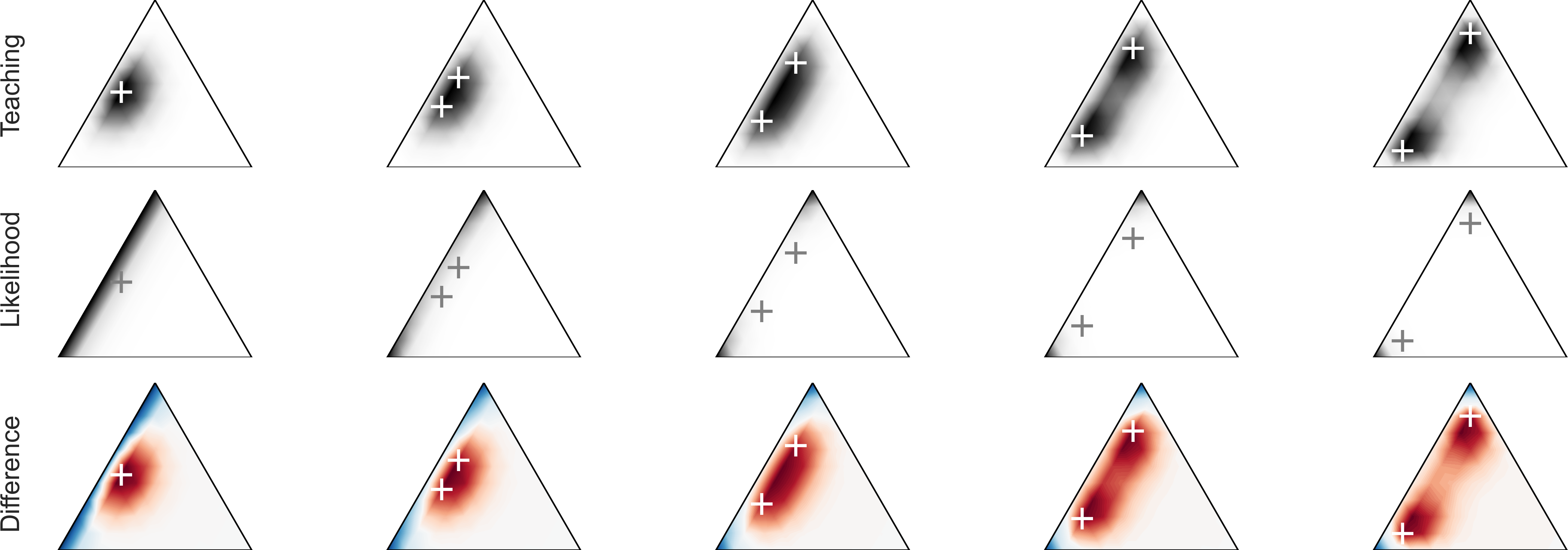}
    \caption{Comparison of the teaching and likelihood distribution representing the probability of selecting a single 10-word document for a two-topic, three-word-vocabulary model where $\alpha = \beta = 0.1$. The 3-dimensional density is represented with barycenter coordinates, where each corner of the triangle represents a word. The two crosses on each plot represent the position of the two topics. Documents are plotted by normalizing their word counts. (Top) The normalized teaching distribution. Darker areas indicated higher density. (Middle) The normalized likelihood. (Bottom) The difference of the two distributions. Red indicates areas in which the teaching distribution has higher density than the likelihood; blue indicates areas in which the likelihood has higher density that the teaching distribution.}
    \label{fig:simplex}
\end{figure*}

We can calculate \autoref{eqn:teaching-topics} exactly when using a small number of small documents with simple topic models. In \autoref{fig:simplex} we plot the teaching distribution of single, 10-word documents under several two-topic models with three-word vocabularies, and $\alpha = \beta=0.1$.\footnote{For simplicity, we use symmetric Dirichlet parameters where $\alpha = c$ implies that each value in $\alpha$ is $c$.} Documents were plotted by using their normalized counts as barycenter coordinates.\footnote{For example, a document $d = [2, 1, 2, 1, 3, 2, 1, 2, 2, 2]$ with counts $[n_1, n_2, n_3] = [3, 6, 1]$ would have the barycenter coordinates $(0.3, 0.6, 0.1)$.} Documents with word proportions more consistent with a topic are closer to that topic on the simplex. 

\autoref{fig:simplex} shows that the teaching model assigns highest density to documents that are closer to both topics while the likelihood (the standard LDA model) assigns higher density to the documents that are at the corners. Because $\alpha$ is low, documents generated by LDA will mostly contain words generated by one topic; the teaching model suggests that choosing documents between the two topics is better for teaching both topics. Because $\beta$ is low, most of the density in topics is expected to lie in a small proportion of the words; this prevents the teaching model from favoring documents centered between the two topics, as when topics are well-separated \parentext{\autoref{fig:simplex}, top right}.

\begin{figure}[htb!]
    \centering
    \includegraphics[trim={0mm 4mm 0mm 3mm}, clip, width=\columnwidth]{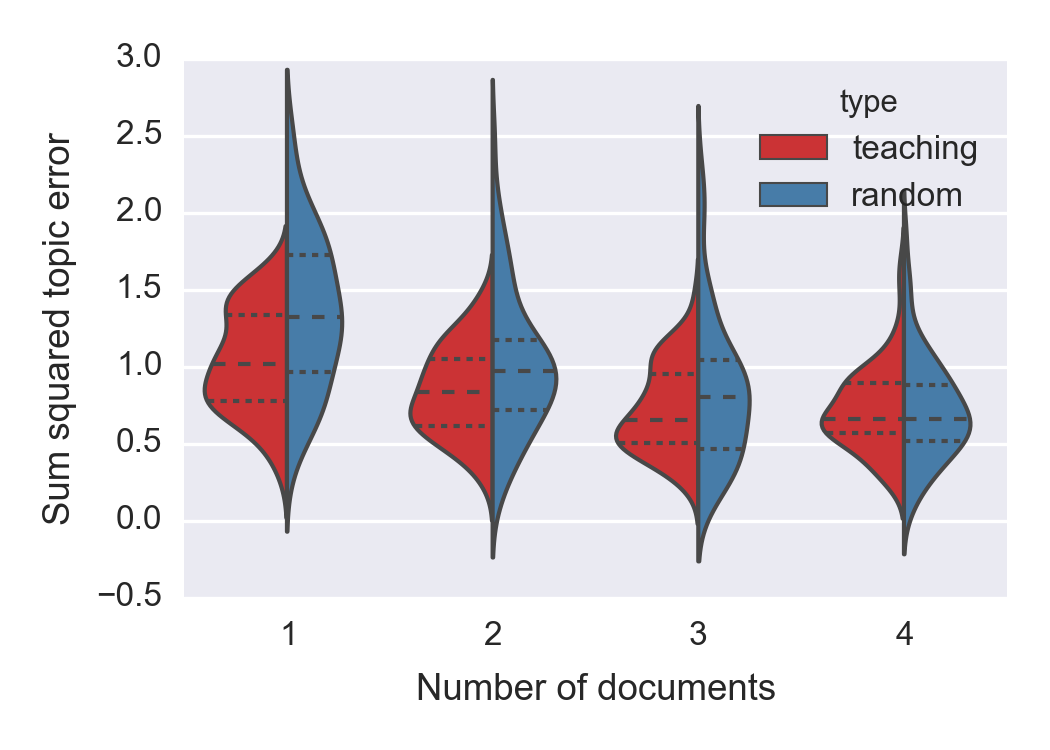}
    \caption{Distribution of error between the inferred and true topics as a function of the number of 20-word documents drawn from the teaching (red) and LDA (random; blue) distributions. Three ten-word topics comprise the true model and $\alpha = \beta = 0.1$. }
    \label{fig:error}
\end{figure}

To determine whether these data benefit learners, we provided LDA with teaching documents and documents drawn randomly from LDA, then computed the error between the inferred topics and the topics used to generate the documents. We generated 64 sets each of one, two, three, and four 20-word documents from the teaching model and from LDA.  We used the pseudo-marginal Metropolis-Hastings algorithm to generate teaching documents. The document sets drawn from LDA served as the starting state for each Markov Chain and proposals were generated by randomly flipping a small number of the words. The likelihood and teaching probabilities were estimated using SIS. We ran LDA \parentext{eqn:lda-gibbs} for 1000 iterations on each document pair. The topics were derived from the counts and compared with the true topics through sum squared error. Due to label switching in the Gibbs sampler, we calculated the error between the true topics and each label permutation of the inferred topics, and report the minimum error across permutations. \autoref{fig:error} shows that the documents from the teaching model produce lower error, and that the effect is greater for fewer documents. As the number of documents increases, the teaching model and random generation produce similar learning outcomes.

The bulk of the benefit of teaching can be expected when learning from fewer documents, but the number of documents at which random and teaching-based sampling are equivalent varies depending on the problem. As the number of topics in the model increase and the the base distribution becomes sparser (as $\alpha$ and $\beta$ approach 0, random documents will contain fewer topics with fewer likely, unique words), the equivalence point will increase. 
In the above example, three teaching documents were beneficial to teach a three-topic model, but four teaching documents produced results similar to documents drawn randomly from LDA. This is consistent with the idea that as the number of documents surpasses the number of topics, there is an increasing probability that random sampling will, by chance, represent all of the topics.
If the the model is not sparse (as $\alpha$ and $\beta$ approach infinity), random documents will contain more topics with more likely, unique words and individual documents will become more similar. When this is the case, relative gains due to teaching will be reduced.

\subsection{Internet Movie Database Top 1000}
\begin{figure*}[t!]
    \begin{subfigure}{\textwidth}
        \centering
        \includegraphics[trim={4mm 4mm 3mm 3mm}, clip, width=\textwidth]{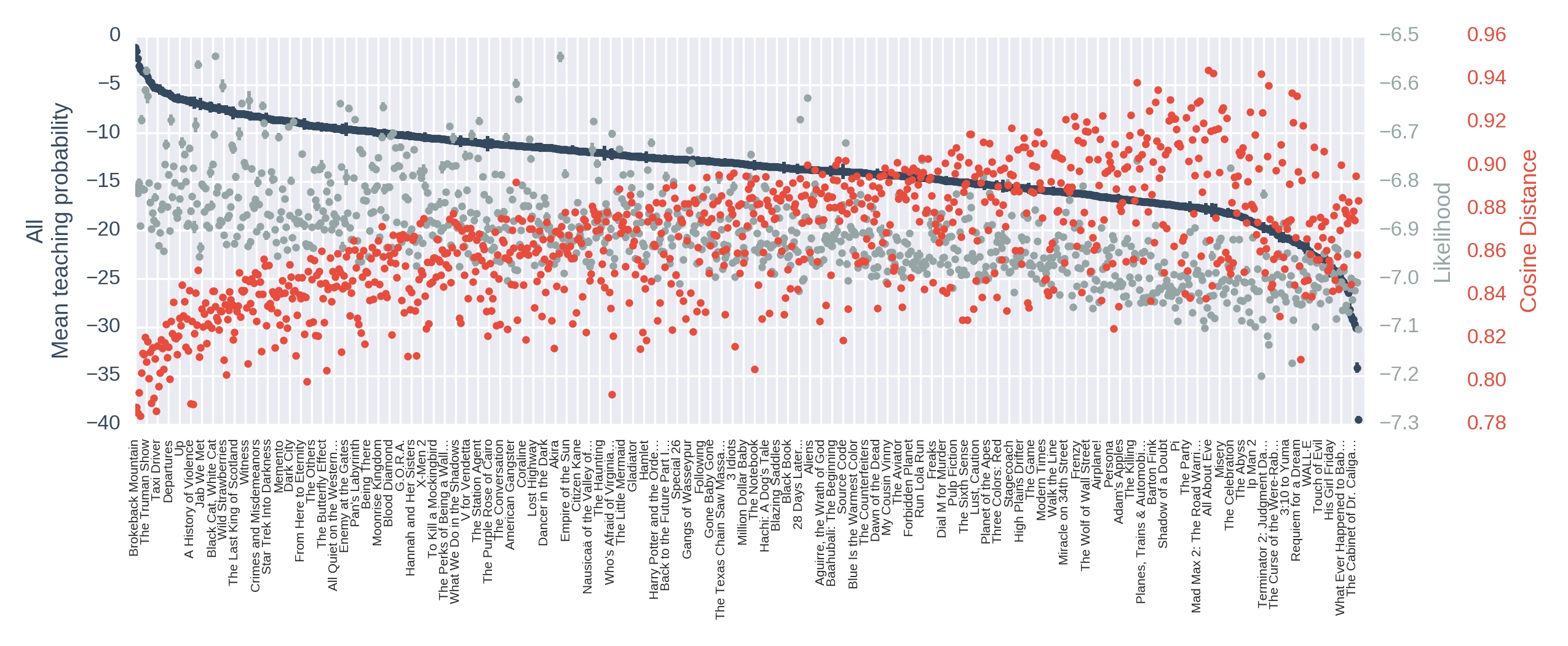}
    \end{subfigure}
    \begin{subfigure}{\textwidth}
        \centering
        \includegraphics[trim={4mm 4mm 3mm 3mm}, clip, width=\textwidth]{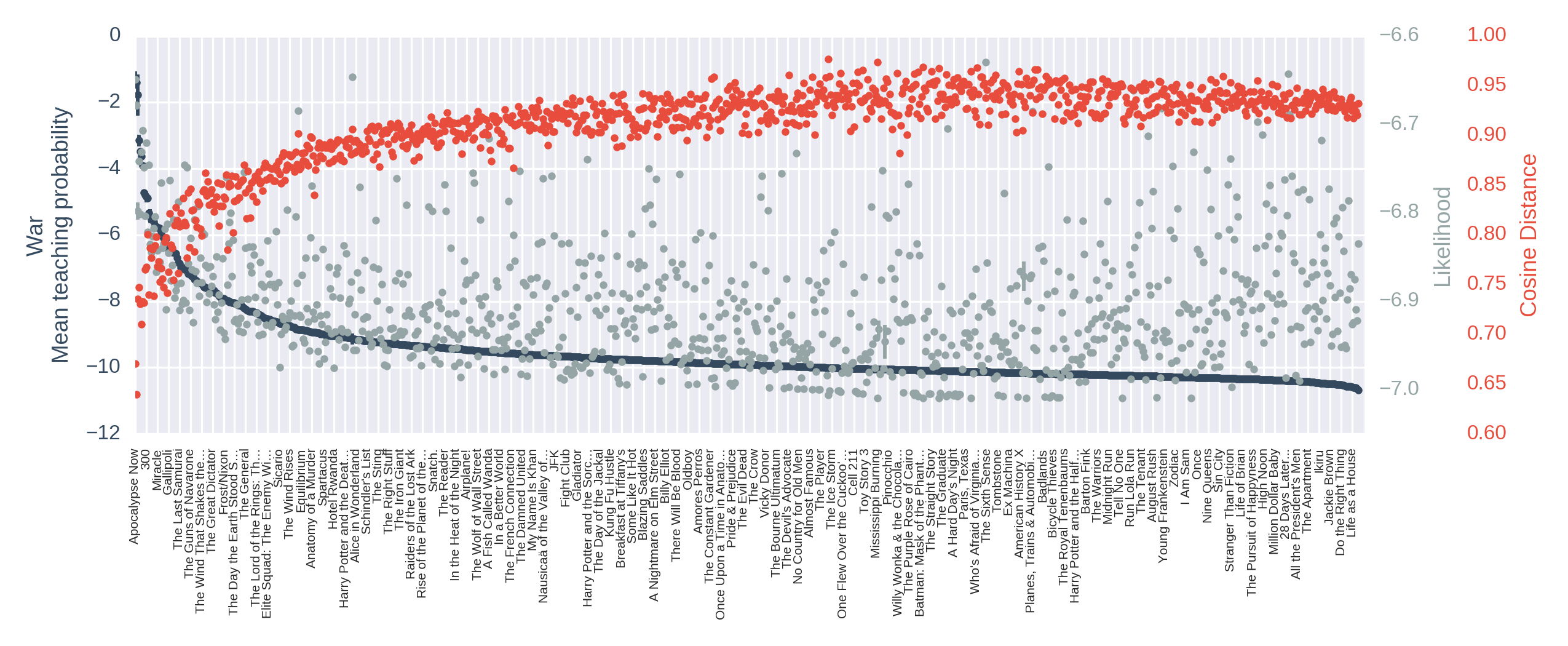}
    \end{subfigure}
    \caption{Comparison of normalized teaching probability (Navy), likelihood adjusted for the number of words in the document ($\mathcal{\ell}^{1/w}$; gray), and one minus cosine between synopses and topic models (red) for the entire topic model (Top) and the \emph{war} topic (Bottom) of The Internet Movie Database top 1000 movies. The leftmost films have the highest probability under the teaching model. Each point represents the mean of sixteen estimates. The standard errors of the estimates are represented, but are often smaller than the points. The title of every 10\textsuperscript{th} film is shown.}
    \label{fig:imdb}
\end{figure*}

To explore the scalability and real-world implications of this work, we applied the teaching model to select movie synopses from the Internet Movie Database top 1000 movies \citep{imdbtop1000}. The synopses were processed in the standard way: stop words
and words that occur fewer than three times were removed, leaving a vocabulary of 3276 words and 41430 total words across 1000 documents. The target topic model, which comprised 16 topics \parentext{the number topics that maximize the evidence for $\alpha=50/T$ and $\beta=.1$, see \textcite{Griffiths2004}}, was derived from running LDA on the synopses. We start by finding the single synopsis that best captures the entire topic model. The teaching probability of each synopsis was calculated 16 independent times using $10^5$ samples from SIS \parentext{see \autoref{fig:imdb}, top}.

The best documents are those that represent the most topics. The best teaching synopsis, \emph{Brokeback Mountain} \citep{Lee2005Brokeback}, is longer than most (64 words) and represents many of the topics---most of which contain drama keywords---though many of the words come from two topics having to do with working and friends/family going on trips. Drawing an analogy to \autoref{fig:simplex}, the ideal single documents lie between topics, but, in keeping with LDA's sparsity, are closer to one than the others; \emph{Brokeback Mountain} exhibits these qualities. 

\autoref{fig:imdb} shows that the likelihood and teaching probabilities correlate to some extent, but that there are substantial differences. Much of the benefit of the teaching probability comes from considering all possible inferences (via the marginal likelihood) and directing learners' inferences toward the target, but away from confusable alternatives.

In some circumstances, one may be interested in teaching subsets of topics $\Phi^* = \{\phi_i^*, \dots, \phi_K^*\} \subset \Phi$. This can be accomplished by marginalizing over the remaining topics. The teacher would like to induce the target topics in the learner but does not explicitly consider the other topics. The numerator in \autoref{eqn:teaching-topics} becomes,
\begin{multline}
    \label{eqn:teaching-subsets}
    \prod_{\phi^* \in \Phi^*} f(\phi^* \mid \beta)
        \sum_{\mathbf{z} \in \mathfrak{Z}} \Bigg(
            \prod_{k=1}^{K}
                \text{DirCat}(z_k \mid \alpha)
            \\ \times
            \prod_{t:\phi_t \not \in \Phi^*}
                \text{DirCat}(w_t \mid \beta)
            \prod_{i}^{n} 
                f(w_i \mid \phi_{z_i})\mathbbm{1}_{\Phi^*}(\phi_{z_i})
        \Bigg),
\end{multline}
where $w_t$ is the set of words assigned to topic $t$.

To demonstrate this feature, we chose to teach a single topic that roughly corresponded to \emph{war}. Some of the most-occurring words in the topic were \emph{war}, \emph{American}, \emph{team}, \emph{army}, \emph{us/US}, \emph{battle}, \emph{mission}, \emph{men}, and \emph{British}. We repeated the procedure above, but used \autoref{eqn:teaching-subsets} in place of \autoref{eqn:teaching-topics}. \emph{Apocalypse Now} \citep{Coppola1979}, which sits atop IMDB's \emph{Top 20 Greatest War Movies of All Time} \citep{Imdbwar} is the best synopsis for teaching the \emph{war} topic. The results \parentext{\autoref{fig:imdb}, bottom} again show that the teaching and likelihood distributions differ significantly. In fact, the two max likelihood synopses, \emph{The Celebration} \citep{Vinterberg1998} and \emph{Mr. Nobody} \citep{VanDormael2009} are both dramas that have nothing to do with war.

To explore the spatial qualities of the teaching distribution, we computed one minus the cosine distance between each synopsis and the normalized sum of the target topics \parentext{\autoref{fig:imdb}, red}. \footnote{One minus the cosine distance between two vectors, $\mathbf{A}$ and $\mathbf{B}$, is 0 when $\mathbf{A}$ and $\mathbf{B}$ are identical.} There is a negative correlation between the cosine distance and the teaching probabilities both for documents under the full topic model ($r(998) = -0.69,\;p\approx 0$) and the \emph{war} topic ($r(998) = -0.93,\;p\approx 0$). 
This result suggests that while selecting documents based on cosine distance is an effective heuristic for approximating teaching the war topic, the teaching model selects documents that capture a richer structure than this simpler approach. When searching for the max teaching probability documents, one may employ cosine distance to reduce the number of teaching probability calculations by computing the teaching probability for only low cosine distance documents. 

\section{Discussion}

The problem of optimally selecting examples for teaching is important across a variety of domains. A general yet scalable method has been elusive because teaching requires simulating the learner. Whereas probabilistic learning can proceed by drawing samples from the posterior distribution, teaching requires approximating the distribution itself. Building from advances in approximate inference, including pseudo-marginal sampling and sequential importance sampling, we introduced a general-purpose approach that provides an accurate simulation-based approximation of optimal teaching examples, and demonstrate by selecting documents to teach the distribution of topics from The Internet Movie Database top 1000 movies \citep{imdbtop1000}. For this problem, simulations suggests that Bayesian teaching scales linearly with the number of words in the teaching document(s). Thought we applied our approach to a problem much larger than could be managed by existing general approaches to teaching, scaling to problems on the order of teaching by selecting books, has yet to be realized. We are optimistic about continued progress toward truly scalable and practically realizable applications.


\printbibliography[title=References]

\end{document}